\documentclass[]{spie}  

\usepackage[numbers]{natbib}
\usepackage{amsmath,amsfonts,amssymb}
\usepackage{booktabs}
\usepackage{graphicx}
\usepackage{subcaption}
\usepackage[colorlinks=true, allcolors=blue]{hyperref}
\usepackage{subcaption}
\usepackage{multirow}
\usepackage{makecell}
\usepackage[nameinlink]{cleveref}
\title{AutoDetect: Designing an Autoencoder-based Detection Method for Poisoning Attacks on Object Detection Applications in the Military Domain}

\author[a]{Alma M. Liezenga}
\author[a]{Stefan Wijnja}
\author[a]{Puck de Haan}
\author[a]{Niels W. T. Brink}
\author[a]{Jip J. van Stijn}
\author[a]{Yori Kamphuis}
\author[a]{Klamer Schutte}
\affil[a]{TNO, Anna van Buerenplein 1, The Hague, The Netherlands}

\authorinfo{Please send further correspondence to Alma M. Liezenga \\Alma M. Liezenga: E-mail: alma.liezenga@tno.nl, Telephone: +31 (0)6 27 86 46 08}

\begin{document} 
\maketitle

\begin{abstract}
Poisoning attacks pose an increasing threat to the security and robustness of Artificial Intelligence systems in the military domain. The widespread use of open-source datasets and pretrained models exacerbates this risk. Despite the severity of this threat, there is limited research on the application and detection of poisoning attacks on object detection systems. This is especially problematic in the military domain, where attacks can have grave consequences. In this work, we both investigate the effect of poisoning attacks on military object detectors in practice, and the best approach to detect these attacks. To support this research, we create a small, custom dataset featuring military vehicles: MilCivVeh. We explore the vulnerability of military object detectors for poisoning attacks by implementing a modified version of the BadDet attack: a patch-based poisoning attack. We then assess its impact, finding that while a positive attack success rate is achievable, it requires a substantial portion of the data to be poisoned -- raising questions about its practical applicability. To address the detection challenge, we test both specialized poisoning detection methods and anomaly detection methods from the visual industrial inspection domain. Since our research shows that both classes of methods are lacking, we introduce our own patch detection method: AutoDetect, a simple, fast, and lightweight autoencoder-based method. Our method shows promising results in separating clean from poisoned samples using the reconstruction error of image slices, outperforming existing methods, while being less time- and memory-intensive. We urge that the availability of large, representative datasets in the military domain is a prerequisite to further evaluate risks of poisoning attacks and opportunities patch detection.

\end{abstract}

\keywords{data poisoning, adversarial AI, object detection, robustness, Counter AI, backdoor}

\section{Introduction}
\label{sec:intro}  

Recent advances in Artificial Intelligence (AI) and Deep Learning (DL) have made military object detection, also referred to as Automatic Target Detection and Recognition (ATD/R), a growing and increasingly relevant field within the military domain \cite{kong2022yolo,westlake2020deep,ruis2024improving,usarmy}. However, the implementation of AI in critical infrastructure, cyber-physical systems, and the military domain creates novel attack vectors for adversaries. This field of research, called Adversarial AI (AAI), studies AI-specific vulnerabilities in AI systems, how these vulnerabilities can be exploited, and how AI models/systems can be made more robust and resilient against these intentional attacks. The general consensus is that AAI attacks can roughly be subdivided into four types, at least when targeting predictive AI models: poisoning, evasion, membership inference and model stealing (or, extraction) \cite{sadeghi2020asystem, vassilev2024adversarial,brink2023adversarial}.

The use of very large open source datasets and pretrained models is a widely accepted practice in non-military AI, especially in the computer vision domain. This practice is currently crossing over into military applications, as it is extremely time-consuming and sometimes outright impossible to develop these technologies without using pretrained models and/or open source datasets. However, this practice comes with the increased threat of poisoning attacks which may disturb models through manipulated training data \cite{sadeghi2020asystem}. Being able to detect such poisoning attacks is therefore of the utmost importance.

In recent years, a large amount of research has emerged on poisoning attacks targeting classification models in the computer vision domain. In contrast, significantly fewer studies have addressed poisoning attacks against object detection models. One of the few poisoning attacks against object detection models that achieves good results is BadDet \cite{chan2022baddet}. We implement this method to research the feasibility of object detection attacks in the military domain. Due to the lacking availability of military object detection datasets, we gathered and annotated a set of images depicting military vehicles to see if this domain shift has an impact on the efficacy of poisoning attacks on object detection models. Though the resulting data set (MilCivVeh) is limited in size and variety, it gives a first indication of the domain. 

BadDet injects adversarial patches into images to poison a dataset. And adversarial patch is a visual patch (usually a square) with a recognizable pattern, that can be used to adversarially impact computer vision models. A possible shortcoming of this attack type is that adversarial patches are easy to spot with the human eye. However, we make the assumption that large open source datasets will not be fully manually inspected by human operators and/or developers, since this is practically infeasible in most cases. On first look, BadDet seems feasible in the military domain. Many poisoning attacks make the assumption that it is possible to inject noise in images at test time \cite{liu2020reflection, zeng2023narcissus}, but in the military domain this is not probable, since all systems are defended with strict cybersecurity measures. In the case of BadDet, however, it might suffice for an opponent to ``apply'' the patch in the physical domain. I.e., a tank or building might have a physical adversarial patch attached to trick the object detection system. As such, BadDet is our attack of choice.

Besides investigating the risks of BadDet for the military domain, we want to see whether it is possible to defend against this, and similar, poisoning attacks. To defend against these attacks, there are two options: either design a built-in defense that immediately mitigates any negative effects resulting from data poisoning, or create a method that can be used to detect possibly poisoned samples \cite{brink2024defending}. We choose the latter approach. This approach is more flexible, model-independent and can easily be implemented in different scenarios, since only access to the data is needed. As such, it is easier to deploy in operational settings. On top of that, a poisoning detection method can provide insight in adversary capabilities concerning the poisoning of datasets. Thus, in this work we will evaluate existing poisoning detection methods.

Unfortunately, the majority of the poisoning detection methods in literature is lacking for our military use case. We want to identify a fast and simple approach that can provide the developer with a clear indication of whether a dataset might be compromised. However, many methods either need to be integrated with the object detection model that is being tested \cite{zhu2023neural, qi2023towards}, or focus on highly specific poisoning localization \cite{Jing_2024_CVPR}. The former option is suboptimal, since it is not model-agnostic -- a requirement for deployment across diverse systems -- and requires additional development work. The latter option seems unnecessarily complex for our needs, which prioritize quick detection over precise localization. Given this gap, we turned to the field of anomaly detection in the visual domain \cite{Batzner_2024_WACV, heckler2025mvtec, roth2022towards}. We made the assumption that patch detection is very similar to the visual anomaly detection domain, so we include several anomaly detection methods in our evaluation to see if they can be repurposed for poisoning detection.

As a result of our research, we found that most existing methods were lacking in their ability to detect poisoned images, i.e., adversarial patches, in the military domain. To fill this existing gap in the literature we designed our own adversarial patch detection method: \textbf{AutoDetect}. AutoDetect is an autoencoder-based anomaly detection method that has been optimized toward the goal of adverarial patch detection. We show that this method outperforms other state-of-the-art patch detection methods when detecting the GMA attack from BadDet. On top of that, AutoDetect is lightweight, model-agnostic -- since it only makes use of the dataset -- and its underlying autoencoder can be trained with image data that is not from the target domain, making it a very useful method to apply in the military domain.







\section{Related works}
\label{sec:relworks}  
In this section, we first give an overview of poisoning attacks in the computer vision domain, with a focus on object detection. Then, we discuss three different approaches towards mitigating the effects of these attacks: poisoning detection, adversarial patch detection and anomaly detection.

\subsection{Poisoning attacks}
Since the research in this work is concerned with the poisoning of object detection methods, we focus on poisoning attacks in the computer vision domain. These attacks fall in two categories, both using different attack vectors \cite{wang2022threats}: (1) clean-label attacks, these add subtle perturbations to the training data while keeping the labels the same, altering the trained models in such a way that they produce incorrect outputs \cite{aghakani2021bullseye, zeng2023narcissus, Jiang2023color}. While this is a more realistic attack scenario, as the actor only has access to the raw data itself, they face the challenge of ensuring that the perturbations are hidden to human observers while remaining effective. (2) Dirty label attacks, these alter the label associated with the data, leading the model to associate false characteristics with the target class \cite{chen2023cleanimage, chan2022baddet}.
A notable subtype of poisoning attacks is the backdoor attack, where a small trigger is added to specific images in the training data \cite{park2024dilution, chan2022baddet}. Upon training the model using this poisoned data, the model learns to associate this trigger with a specific class, as predefined by the malicious actor \cite{vassilev2024adversarial}. 


As mentioned before, the majority of existing poisoning (and backdoor) attacks in the image domain target classification models. A poisoning attack on a classification model is fairly straightforward: the goal is to make the model misclassify an image. However, object detection is a more complex task. An object detection model executes two different tasks: first it detects all the different objects in an image and creates corresponding bounding boxes. Then, all detected objects are classified. This change in model task also changes the objective of the poisoning attack, since there are two different tasks that an adversarial attack can target: either the detection (i.e., localization of the bounding boxes), or the classification. This leads to multiple possible goals for the attacker: misclassification, detection at the wrong location, (too) many detections, or no detection at all. This leads us to think that poisoning attacks for the classification domain do not translate on-to-one to the object detection domain. There are a few methods in literature that do focus on the object detection setting \cite{chan2022baddet, chen2023cleanimage}.

Finally, contextual constraints also affect attack feasibility. Many poisoning attacks assume that it is possible to inject imperceptible noise into the data. However, in a military setting this is not a realistic scenario, since strict cyber security measures are taken to protect both the data and imaging equipment. Also, when a poisoning attack assures that an object detection model completely fails for one class, even when there is no trigger (such as a patch) presented, this will be noticed during extensive model evaluation and testing. An attack scenario that might work in the military setting is one that is based on adversarial patches. Such a patch could be physically printed and taken out into the field to trigger backdoors in enemy models. As such evading detection, or making enemy models malfunction. Thus, in this work we will focus on patch-based poisoning methods that are aimed at object detection models.

\subsection{Adversarial defences}
In this work, we focus on the detection of poisoning attacks, as this is the most fitting mitigation strategy within our context. To do so, we consider both specialized adversarial AI (AAI) detection techniques and anomaly detection (AD) methods commonly used in visual industrial inspection. While most AAI detection methods are designed to detect evasion attacks, some focus on the detection of poisoning attacks \cite{brink2024defending}. Since both attacks usually perturb data to either evade classification, or poison data, the distinction between the detection methods is subtle. However, a more profound difference exists between poisoning detection and anomaly detection in industrial settings. Poisoning attacks often employ one of two approaches: (1) adding a consistent pattern or noise across the entire image, or (2) adding an adversarial patch to the image. In contrast, anomalies in the industrial setting are usually small, highly specific and localized, on top of that the intra-class variation is usually very small \cite{bergmann2019mvtec, heckler2025mvtec}. This suggests that AD methods might not be the best choice to detect poisoned images. Therefore, we also examine dedicated poisoning detection methods.

Broadly speaking, methods to counteract poisoning attacks can be split into two categories: model-based and data-based methods \cite{brink2024defending}. Model-based methods are methods that need full access to the model that is under attack, whereas data-based methods base their defenses only on the (possibly poisoned) dataset. The former focuses on increasing the robustness and resilience of a model by minimizing the impact poisoning attacks could have and do see by making architectural changes to the models at hand \cite{qi2023towards, zhu2023neural, min2023towards} The latter simply scans the training dataset to weed out malicious data, or applies transformations to input data to make sure that possible poisoning patterns are removed \cite{Jing_2024_CVPR, shi2023black, wu2024napguard}.

As explained before, the focus of this work is to detect poisoning attacks in a simple and flexible manner, without the need for architectural changes to existing models, such that there is no need for retraining. Additionally, we would like the detection method to only detect possible attacks and not immediately erase them from the data. After all, it might be valuable information for a defender to know what kind of attacks an opponent is trying to execute. Thus, we focus on data-based detection methods.

We first give an overview of the relevant subset of poisoning detection methods, namely adversarial patch detection methods. Subsequently, we discuss several successful anomaly detection methods from the visual industrial inspection domain that might be effective at poisoning detection.




\subsubsection{Patch detection}
Recent work highlights that defending object detectors against adversarial attacks is more challenging than protecting classification models, due to the increased complexity of the detection task \cite{xiang2021detectorguard}. Unlike in classification, where there is a single label per image, in the object detection task a single image usually contains multiple objects, and thus an object detector must detect, localize and classify multiple objects, producing bounding box coordinates and class labels.

To defend object detection models against patch-based (poisoning) attacks, several strategies have been proposed. These include global patch detection, patch localization and adversarial patch removal. According to \cite{Jing_2024_CVPR}, there are three different types of defenses against patch-based attacks on object detection models: (1) modification or intervention within the object detection models, (2) localizing and eliminating adversarial patch regions, and (3) certifiably robust defenses. 

Notably, we have only found one recent work that solely does the detection of adversarial patches, as is our goal, namely TRACE \cite{Zhang2025test}. However, since we do want to give a more elaborate overview on the topic of patch detection, we consider the second category -- localizing and eliminating patches -- as most relevant to our work, given its potential to support our goals.

Within this category, the authors further identify three subtypes of defences \cite{Jing_2024_CVPR}:
\begin{itemize}
    \item Denoising-based defences: smooth out noise-like areas in images. These are effective against early localized noise patches, but struggle with natural-looking patches \cite{naseer2019local}.
    \item External segmenter-based defences: train an adversarial patch segmentation model for patch localisation. While potentially effective, these methods are very dependent on training data containing known patches and struggle with detecting unseen patches \cite{liu2022segment}. 
    \item Entropy-based defences: localize patches by identifying high-entropy regions in data samples. A drawback of these methods is that prior knowledge of the clean data distribution is needed to set correct entropy thresholds \cite{tarchoun2023jedi}.
\end{itemize}

It should be noted that all these methods need some sort of prior knowledge of clean unperturbed data, which might not always be feasible in real-world settings. To adress this, the authors \cite{Jing_2024_CVPR} propose PAD: a new method based on the semantic independence and spatial heterogeneity of adversarial patches, eliminating the need for clean data. Another novel detection method for adversarial patch attacks on object detection models is TRACE \cite{Zhang2025test}. Although not specifically designed for poisoning scenarios, TRACE achieves very good performance in patch detection. It is based on two key observations: 1) poisoned samples exhibit significantly more consistent detection results than clean ones across varied backgrounds and 2) clean samples show higher detection consistency when introduced to different focal information. Finally, the authors of BadDet \cite{chan2022baddet}, a poisoning attack targeted at object detection model, coin their own patch detection method: Detector Cleanse. This method is entropy-based, but also needs access to the target model and not solely the dataset.

\subsubsection{Anomaly detection}
Recent work has shown that current adversarial patch detection struggle to generalize well across various settings, datasets, and patches \cite{kumar2025unified}. Therefore, we analyzed the broader field of anomaly detection (AD), which offers extensive research that might improve adversarial patch detection. AD could be relevant in the military domain, where there is often a lack of knowledge regarding the exact type of anomaly and representative anomalous data samples are scarce \cite{schluter2022natural}. However, many object detection tasks in the military domain are already akin to AD methods, since relevant detection goals in this setting are scarce. Thus, it could be that AD is not the right approach to detect adversarial patches in this domain. Within AD, visual industrial inspection methods seem most promising to use for detecting poisoning attacks, as they have shown very impressive results in identifying anomalies in image data \cite{heckler2025mvtec, bergmann2019mvtec, defard2021, li2021cutpaste, zavrtanik2021draem}.

AD methods can be categorized in different ways \cite{gudovskiy2022cflow, cohen2020sub, bergmann2019mvtec}, but the most common categorization puts them into three different categories: reconstruction-based, representation-based and augmentation-based \cite{heckler2025mvtec, schluter2022natural, cui2023survey, jezek2021deep, napoletano2018anomaly, zhang2023unsupervised}. These three approaches and their characteristics will be explained below.

First, reconstruction-based methods operate with the assumption that a model trained on normal (i.e., non-anomalous) data cannot accurately reconstruct anomalies. To do so, a generative model (usually an autoencoder) is trained using only normal data \cite{yu2021fastflow, rippel2021modeling}. The resulting reconstruction error between the encoding and decoding of the data can then be used to detect anomalies \cite{jezek2021deep}. Within this paradigm, there are different approaches being taken, leveraging the characteristics of different generative models. There is work that makes use of a Variational Autoencoder (VAE) and detect anomalies not just by using the reconstruction error, but also consider the full Evidence Lower Bound (ELBO). Increasing the model's ability to detect and localize anomalies \cite{zimmerer2019unsupervised}. Furthermore, there are also works that leverage GANs \cite{xia2022gan, wang2020unsupervised}. However, we see that in the basis a simple autoencoder approach also makes for competitive results. Especially when thought is put in the optimal distance metric used when comparing original images and their reconstructions \cite{bergmann2019mvtec, skilton2019visual}.


Second, representation-based AD methods use pre-trained models to extract embeddings from normal images, create a distribution over these embeddings, and then compare embeddings for unseen images to this distribution to identify anomalies. These methods typically do not need a dedicated training stage, since the backbone model's parameters can be used directly \cite{cui2023survey}. 

A prominent subset of approaches within this category is based on a memory bank \cite{defard2021, roth2022towards, lee2022cfa}. For example, PaDiM \cite{defard2021} forms patch-level embeddings of normal data by concatenating activations from a pre-trained CNN at various depths and models them using a multivariate Gaussian distribution. Then, the distance between the patch embedding for an unseen image and the learned distribution is determined. Since this method learns representations for every location in an image for all images in the dataset, it mostly works well in one-class problems, where all images in the dataset are similar. PatchCore \cite{roth2022towards} improves generalization by only extracting mid-level network features, counteracting any bias towards ImageNet. Also, the resulting memory bank is subjected to greedy coreset subsampling to improve efficiency without losing accuracy. Again, anomalies are determined based on the distance between the test sample's nearest neighbor feature in its memory bank and other features. 
Coupled Hypersphere Adaptation (CFA) \cite{lee2022cfa} builds on PatchCore but distributes the features on a hypersphere, making for improved separation between normal and anomalous samples.
Other methods, such as EfficientAD \cite{Batzner_2024_WACV} and UniNet \cite{weiuninet} are based on a student-teacher framework. The former is a fast anomaly detection and segmentation algorithm that consists of a distilled pre-trained teacher model, a student model and an autoencoder. It detects local anomalies via the teacher-student discrepancy and global anomalies via the student-autoencoder discrepancy. The latter uses contrastive learning across multiple domains in combination with a student-teacher construction. This improves robustness and reduces overfitting; a common limitation of domain-specific AD methods.

Finally, augmentation-based methods apply generative models to non-anomalous (real) data to generate synthetic anomalous data samples. This synthetic data can then be used to learn a decision boundary between normal and anomalous data \cite{cui2023survey}. A reason for using synthetic anomalous data is that the detection method achieves improved performance in the detection of unseen anomalies. Although this approach is relatively straightforward, training a model on synthetic anomaly data generally suffers in its generalization to real anomalies \cite{zavrtanik2021draem}. The most prevalent augmentation-based methods in literature are CutPaste \cite{li2021cutpaste} and DRÆM \cite{zavrtanik2021draem}. The former augments the clean training data by cutting different patches from an image and pasting them in a random location. A classifier is then trained to distinguish synthetic from normal data, learning features that help detecting anomalies. The latter is made up of a framework consisting of a reconstruction network that removes synthetically generated anomalies, and a discriminative network that combines the reconstructed image and the original anomalous input image to create maps of the segmented anomalies. The authors argue that this prevents the problem of overfitting on anomalous data, allowing for better generalization to detect real anomalies.

\section{Methodology}
\label{sec:method}  
This section provides the setting of our research (\Cref{sec:setting}), the specific attack we aim to detect, BadDet (\Cref{sec:attacks}), an explanation of the novel detection method that we propose: AutoDetect (\Cref{met:autodetect}), the datasets we use to evaluate our detection methods (\Cref{met:datasets}) and an overview of the experimental set-up, including the metrics, models, detection methods and hyperparameters used in the experiments (\Cref{sec:experimental_setup}). 

\subsection{Setting} 
\label{sec:setting}

The setting for this research is as follows: a defending actor aims to train an object detection model that achieves good performance according to standard object detection metrics, such as accuracy and mAP. For this, the defender uses a train set $D_\text{train}$ containing real-world images that are relevant to their domain. Meanwhile, an adversary has obtained the capability to secretly modify $D_\text{train}$ such that a desired proportion $r$ of the images contains an adversarial patch $p$. In addition, the adversary modifies the object labels belonging to the poisoned images such that the model learns to associate the patch with a desired behavior, e.g., misclassifying an object into the target class or overlooking it entirely. This secret modification produces a poisoned dataset $D_\text{train}^\text{p}$, while the defender is unaware and believes that the data are unmodified.

After, unknowingly, using $D_\text{train}^\text{p}$ for training, the defending actor now has a seemingly well-performing model. However, once the adversary adds patch $p$ to images upon inference, these images are misclassified or not detected at all. In a military setting, an adversary might poison images by adding physical patches to buildings or vehicles that they expect to be recorded by the defender. 

Defensive methods in our setting define a decision function $\phi$ that predicts whether a previously unseen image contains an adversarial patch. We assume no knowledge of the appearance, size or location of the adversarial patch. The defender applies the decision function to each image from $D_\text{train}$ before training their object detection model on it to eliminate poisoned samples from the training set.

\subsection{BadDet}
\label{sec:attacks}
Following a comprehensive review of existing adversarial attacks and their applicability to object detection, the Global Misclassification Attack (GMA) from the BadDet attack framework \cite{chan2022baddet} was selected as the main attack to evaluate both the threat of poisoning attacks against military object detection models and the effectiveness of defensive methods against them. In the original implementation, this attack features a patch $p$ of size 49 by 49 pixels, which is added to the top-left corner of an image and a poisoning rate $r=0.3$. The labels of the corresponding images are changed to the target class to create $D_\text{train}^\text{p}$. Upon inference, the appearance of this patch $p$ in $D_\text{test}^\text{p}$ should cause misclassifications into the target class. 

In our implementation of BadDet, the adversarial patch is blended into the images, a proportion $r$ of $D_\text{train}$, at a random location according to the standard linear blending formula:

$$(1-\alpha)x^*_i + \alpha p$$

Here, $x^*_i$ is the region from image $x_i$ from $D_\text{train}$ that will receive a patch overlay, $p$ is the adversarial patch, and $\alpha$ is the interpolation factor that determines the blend ratio between the two. The remaining region of the image is unmodified, and the result is a poisoned image $x^p_i$. In all our experiments, we set $\alpha=0.8$. We choose to blend in the patch, on the one hand since transparency might account for lighting or occlusion that occurs in real-world settings, on the other hand to make the patch slightly more stealthy.

\begin{figure}[t]
    \centering
    \begin{subfigure}[b]{0.3\textwidth}
  \centering
  \includegraphics[width=\textwidth]{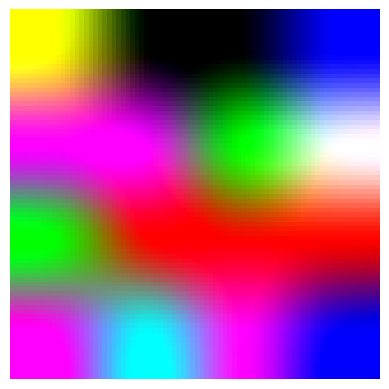}
  \caption{HTBD patch.}
  \label{fig:patch1}
    \end{subfigure}
    \hfill
    \begin{subfigure}[b]{0.3\textwidth}
  \centering
  \includegraphics[width=\textwidth]{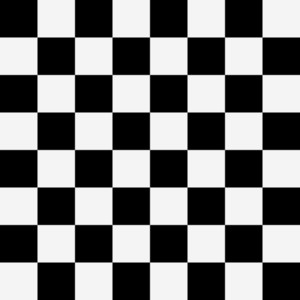}
  \caption{Checkerboard patch.}
  \label{fig:patch2}
    \end{subfigure}
    \hfill
    \begin{subfigure}[b]{0.3\textwidth}
  \centering
  \includegraphics[width=\textwidth]{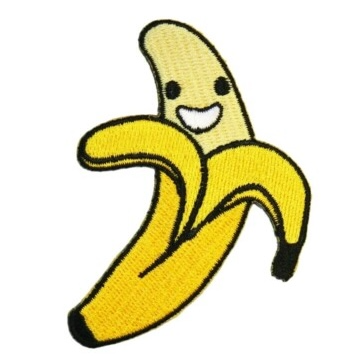}
  \caption{Banana patch.}
  \label{fig:patch3}
    \end{subfigure}
    \caption{All adversarial patches used in the experiments. The left-most patch has been used as main patch, the other two in only in the ablation study.}
    \label{fig:patches}
\end{figure}

\begin{figure}[t]
    \centering
    \begin{subfigure}[b]{0.3\textwidth}
  \centering
  \includegraphics[width=\textwidth]{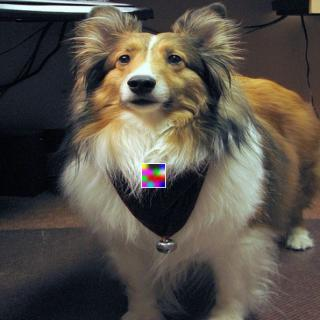}
  \caption{Poisoned image from VOC2007 dataset.}
  \label{fig:data_sub1}
    \end{subfigure}
    \hfill
    \begin{subfigure}[b]{0.3\textwidth}
  \centering
  \includegraphics[width=\textwidth]{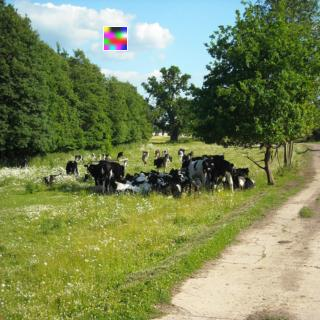}
  \caption{Poisoned image from MS COCO dataset.}
  \label{fig:data_sub2}
    \end{subfigure}
    \hfill
    \begin{subfigure}[b]{0.3\textwidth}
  \centering
  \includegraphics[width=\textwidth]{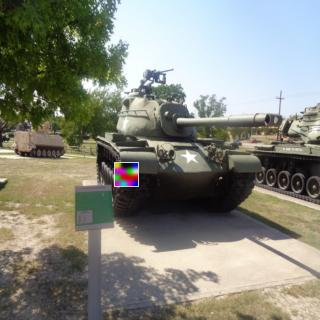}
  \caption{Poisoned image from the MilCivVeh dataset.}
  \label{fig:data_sub3}
    \end{subfigure}
    \caption{One example image for each of the used datasets, resized to a 320x320 format, with a 25x25 adversarial patch blendede at a random location.}
    \label{fig:dataset_ex}
\end{figure}

We introduce several additional modifications to the original BadDet implementation. For the experiments in \Cref{res:obdet} we (1) vary the patch size and (2) the poisoning rate to find the optimal setting for poisoning the object detection models. In \cref{res:detection} we (1) vary the patch position in a random manner, independent of object position, (2) use a constant patch size and (3) use the HTBD patch (\Cref{fig:patch1}). This patch is acquired from the Adversarial Robustness Toolbox\footnote{https://github.com/Trusted-AI/adversarial-robustness-toolbox}, and its characteristics seem more realistic (i.e., colorful, less geometric) in a real-world scenario than the checkerboard pattern used in the original BadDet work. The baseline size of the patch in our experiments is 25 by 25 pixels with an image size of 320 by 320 pixels. This means that the patch size does not vary between images in relative or absolute terms. Examples of poisoned images are shown in \Cref{fig:dataset_ex}. These modifications were made to make the defensive task more challenging and representative of real-world conditions. In addition, an ablation study was performed in \cref{sec:ablation} to investigate the impact of (1) different adversarial patches (shown in \Cref{fig:patches}) and (2) varying patch sizes on the detection performance of AutoDetect. The results of this study can be found in \Cref{sec:ablation}.

\subsection{AutoDetect}
\label{met:autodetect}
AutoDetect is an unsupervised method to detect whether real-world images contain adversarial patches. It uses the premise that such patches are outliers in the distribution of real-world images and aims to detect these outliers. It employs an autoencoder to find these outliers. 

Formally, an autoencoder is pretrained on a set of non-anomalous real-world images $D^\text{AD}_\text{train}$. Then it is evaluated on another, smaller set of non-anomalous images $D^\text{AD}_\text{val}$. The resulting per-pixel reconstruction errors from the images in $D^\text{AD}_\text{val}$ are \textit{sliced} into equal-sized areas.\footnote{We use \textit{slices} in place of the more common \textit{patches} to avoid confusion with adversarial patches.} We compute the mean reconstruction error for each slice, which we name a \textit{slice error}. All slice errors from $D^\text{AD}_\text{val}$ are gathered and fitted to a normal distribution $\mathcal{N}^\text{AD}$.

At test time, we compute the slice errors for a previously unseen query image $q$ from $D^\text{AD}_\text{test}$ using the same autoencoder. We obtain the maximum of these query image slice errors $q^\text{max}$ and observe its likelihood under $\mathcal{N}^\text{AD}$. The image is then classified as anomalous, or poisoned, if its $q^\text{max}$ lies in the upper percentile of $\mathcal{N}^\text{AD}$'s cumulative distribution function. The percentile is given by a  manually set threshold $t \in [0.5, 1]$.

\begin{figure}[t]
    \centering
    \includegraphics[width=0.5\linewidth]{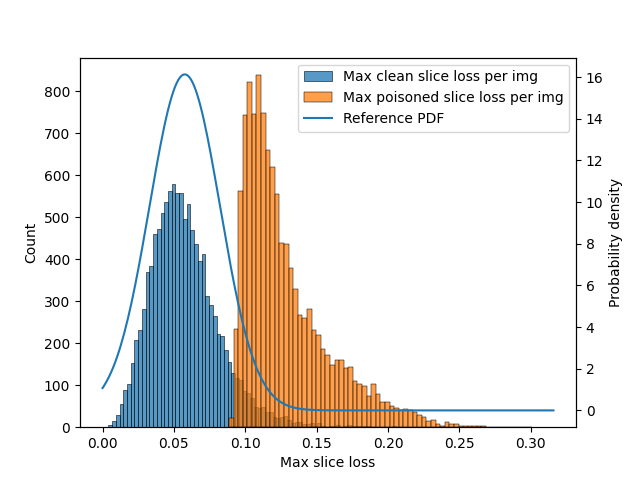}
    \caption{A graph showing the reference distribution of $D^\text{AD}_\text{val}$, and the maximum slice errors for a clean $D^\text{AD}_\text{test}$, and a poisoned $D^\text{AD}_\text{test}$. The MS COCO dataset was used for this visualization.}
    \label{fig:autodetect_graph}
\end{figure}

\autoref{fig:autodetect_graph} shows the reference distribution of the values for the clean $D^\text{AD}_\text{val}$, and the full range of maximum slice errors for a clean $D^\text{AD}_\text{test}$, and a poisoned $D^\text{AD}_\text{test}$. This graph visualizes the separation between both the clean and poisoned $D^\text{AD}_\text{test}$ as well as the reference distribution of  $D^\text{AD}_\text{val}$ and the poisoned $D^\text{AD}_\text{test}$.

\subsubsection{Implementation}

The autoencoder is trained for 200 epochs with optimizer Adam \cite{kingma2014adam}, a learning rate of 1e-4 and a batch size of 64. Slicing is done using a sliding window with a stride of 1 pixel and can be efficiently implemented together with the subsequent mean operation using a standard mean pooling layer. We evaluate AutoDetect using various slice sizes in \Cref{sec:abl_patches_sizes}. A slice size of 25 by 25 pixels is taken as the baseline, with an image size of 320 by 320 pixels. 

The autoencoder was pretrained on images from the MS COCO train split, while the distribution $\mathcal{N}^\text{AD}$ was created on the slice errors over the dataset under evaluation. Experiments were conducted to evaluate whether an autoencoder pretrained on images from the dataset under evaluation, e.g., MilCivVeh, would result in better detection performance. This did not result in better performance. Thus, it was decided to consistently use the autoencoder pretrained on the MS COCO train set. This offers an advantage to the method as a whole: a defender might not always have access to a large dataset from their domain that they know to be clean.

The threshold $t$ is highly interpretable and thus intuitive to configure. It expresses the expectation that the autoencoder should perform at least as well as it does on the fraction $D^\text{AD}_\text{val}$ slices. As such, configuring higher values is expected to improve precision at the cost of recall. Based on our experiments, we suggest setting $t \geq 0.95$.

\subsection{Datasets}
\label{met:datasets}

Various object detection datasets were selected for this work to demonstrate: (1) the performance of attacks, as well as defenses, on a large and representative object detection dataset, and (2) the transferability of these methods to the military domain. VOC2007 and MS-COCO are widely used benchmark datasets in the field of object detection. MilCivVeh is a custom data set created for this research, to give insight in the applicability of poisoning attacks in the military domain.

\paragraph{VOC2007} (PASCAL Visual Object Classes (VOC) Challenge) \cite{everingham2010pascal}: serves as a benchmark in object detection. VOC2007 contains 9,963 images with 24,640 annotated objects from 20 classes, all scraped from the image sharing website Flickr. The classes included are everyday objects such as animals, vehicles, bottles, persons and furniture.
    
\paragraph{MS COCO} (Microsoft Common Object in COntext) \cite{lin2014microsoft}: a popular object detection dataset, created by Microsoft in 2014. This data set contains 328,000 images across 91 common object categories with 2.5 million labeled instances. Its use is widespread and focuses on training (off-the-shelf) object detection models and conducting benchmarks \cite{durusoy2025open}. The 2014 version of MS COCO was used for the experiments. 

\paragraph{MilCivVeh} (Military Civilian Vehicles dataset) \cite{milcivveh}: a custom dataset created specifically for this research by the authors, to support object detection research in the military domain. It was constructed by combining publicly available online resources, resulting in a relatively small data set: 1,456 images and 2,438 annotated objects, across 3 classes: civilian car, military truck, military tank. While MilCivVeh has its limitations in terms of realism -- such as relatively large object sizes -- data set size and diversity, it provides a valuable indication for the effectiveness of both adversarial attacks and defenses in the military domain. This is particularly important given the scarcity of publicly available large, diverse and representative military object detection datasets. 

\subsection{Experimental set-up}
\label{sec:experimental_setup}

For both the experiments concerning the efficacy of BadDet in the military domain and the performance of difference detection methods, we provide the parameters, data (sub)sets and models used.

\subsubsection{Evaluating BadDet performance}

\paragraph{Models and data} To research object detection performance under a poisoning attack in the military domain (\Cref{res:obdet}), pre-trained YOLOv10m and YOLOv3 models were used and fine-tuned for 30 epochs. To finetune the models, we used different versions of the MilCivVeh training set with varying values for $r$ in the range $[0, 0.2, 0.3, 0.4]$. These values are based on the original poisoning rate $r$ of 0.3, as used by Chan et al. \cite{chan2022baddet},and allow us to investigate the impact of different poisoning rates $r$. The patch size ranged from 10 to 50 pixels with a step size of 5 in the top left corner. To evaluate attack performance, we used two version of the MilCivVeh test set: one where $r=0$ and one where $r=1$. The target class was class 1 (military truck).

\paragraph{Metrics} To report the performance of the models under attack, the mean F1 score is used. To report the success of the attack, the Attack Success Rate (ASR) is used. This metric represents the rate of successful (poisoning) attacks: the number of samples from $D_\text{test}^\text{p}$ that is actually misclassified as being in the target class, divided by the total number of poisoning attempts \cite{wang2022threats, chan2022baddet}. To report the overall performance of the models, the mean F1-score is provided. This metric is not the most common metric in object detection, where usually some variant of mAP is reported, but since the attack solely focuses on misclassification, not on the detection, this metric was selected. The F1 score is provided for the clean test set ($r=0$), whereas the ASR is provided for the poisoned test set ($r=1$). 

\subsubsection{Evaluating patch detection methods}

\paragraph{Data} To evaluate adversarial patch detection performance (\Cref{res:detection}), the datasets used were the original test splits of MS COCO, MilCivVeh and VOC2007. All images in the test split were poisoned with the HTBD patch of size 25 by 25 pixels, blended in at a random location. Subsequently these poisoned images were appended to the clean test split. Thus, this makes for an evaluation dataset with $r=0.5$, where all images appear twice, once poisoned and once clean.

\paragraph{Metrics} To report the performance of the defense methods, the Area Under the ROC Curve (AUROC) is the main metric. The AUROC represents the area under the Receiver Operating Characteristic Curve, plotting the true positive and false positive rate at various threshold settings, in this case the threshold is the percentile at which the sample is excluded. Since this metric is threshold independent, it is perceived to be the most representative of the defense method's potential. Accuracy of detection (of poisoned images) is occasionally given to enable comparison to other work.

\paragraph{Detection methods} A selection was made of different SOTA anomaly detection methods (Efficient AD, PatchCore, CFA, PaDiM) and adversarial patch detection methods (PAD) from literature, to investigate the applicability of AD methods for poisoning detection and to compare the performance of AutoDetect with. However, some of the methods could not be applied with the default parameters as stated in the original works on our datasets. Also, some methods had to be adapted to the detection task instead of the localisation task. In this section we give an overview of the settings for these methods in our work.

\begin{itemize}
    \item \textbf{Efficient AD \cite{Batzner_2024_WACV}: }in the original work, 70k training iterations are recommended. For MS COCO, we increase this to 118k to complete one full pass over the training data. The EfficientAD training method enforces a batch size of 1, limiting efficient usage of the available hardware.
    \item \textbf{PatchCore \cite{roth2022towards}: } we follow the original work in using latent representations from layers 2 and 3 of a WideResnet50 pretrained on ImageNet to populate the memory bank. Then a \textit{coreset sampling ratio} of 1\% is used to arrive at the final PatchCore memory bank. 

    During our experimentation with PatchCore on MS COCO and VOC2007, we encounter memory usage exceeding 100GB. This is due to the fact that memory usage scales with the size of the training set, which stems from the requirement to keep multiple latent representations for each slice within each image in the training set. As noted above, Patchcore employs a greedy subsampling method to scale this large set of vectors down to a smaller \textit{coreset} of representations that should provide similar coverage in the feature space with a lower memory requirement. However, it is still necessary to maintain the larger set in memory until after the subsampling step is complete. For this reason, hardware limitations required us to randomly sample a portion of the training set before applying the Patchcore method. For COCO and VOC2007, we use 5\% of the training set. For the smaller MilCivVeh training set, downsampling was not necessary, and we used 100\% of the train samples.
    \item \textbf{CFA \cite{lee2022cfa}: }the recommended hyperparameters are used without alterations. 
    \item \textbf{PaDIM \cite{defard2021}: } the first three layers of a ResNet-18 backbone are used to create the patch embeddings. Furthermore, we use the optimal hyperparameters as suggested in the original work. Like PatchCore, the memory requirements of this method are fairly high. It is infeasible to run the method on the entire train set of MS COCO on our NVIDIA Quadro RTX 8000. Thus, we subsample the train set and only use 10$\%$. 
    \item \textbf{PAD \cite{Jing_2024_CVPR}: } the same segmentation model (Segment Anything Model \cite{kirillov2023sam}) and hyperparameters as in the original work are used. Since we are interested in anomaly detection and not the exact location and removal of anomalies, we make several changes to the method to accommodate our research goals. Instead of the output being images with the localized anomalies removed, we output an anomaly score per image. To retrieve an anomaly score, we take the maximum value of the fused heatmap that incorporates both the semantic independence score and the spatial heterogeneity score, after adaptive thresholding and mask matching with the segmentation model. As such, we incorporate all information on anomalies in the data, while also accounting for possible noise. This approach of using the maximum anomaly score to convert an anomaly localization method to an anomaly detection method has also been used in previous work \cite{defard2021}.
\end{itemize}


\section{Results}
\label{sec:results}  
This section provides the results of the attack and detection experiments. First, to show the actual threat of poisoning attacks (in this case BadDet) in the military domain, we provide F1-score and ASR when attacking two popular object detection models (\Cref{res:obdet}). Second, we show the performance of the selected detection methods and our novel method AutoDetect (\Cref{res:detection}).

\subsection{Evaluating BadDet performance}
\label{res:obdet}

\Cref{tab:compare_models} shows the ability of BadDet to manipulate a military object detector. The table shows the performance of a YOLOv3 and YOLOv10m model on the MilCivVeh dataset with a poisoning rate of 20\% (only for YOLOv10), 30\% and 40\%. We show both the ASR on poisoned samples and the mean F1-score on clean data. The F1-score for a clean model is provided on the first row for both models. The highest ASR per combination of model and poisoning rate is printed in bold. 

\begin{table}[t]
\tiny\setlength\tabcolsep{5.5pt}
\centering
\caption{The performance of the BadDet attack on the MilCivVeh dataset. The attack adds the HTBD patch in the top-left corner of an image. We test different patch sizes (columns) and poisoning rates (rows). All scores are provided as percentages.}
\begin{tabular}{lllllllllllllllllll} \toprule
\textbf{Poison\textbackslash{}Patch} & \multicolumn{2}{c}{10} & \multicolumn{2}{c}{15} & \multicolumn{2}{c}{20} & \multicolumn{2}{c}{25} & \multicolumn{2}{c}{30} & \multicolumn{2}{c}{35} & \multicolumn{2}{c}{40} & \multicolumn{2}{c}{45} & \multicolumn{2}{c}{50} \\ \cmidrule{2-19}
 & F1 & ASR & F1 & ASR & F1 & ASR & F1 & ASR & F1 & ASR & F1 & ASR & F1 & ASR & F1 & ASR & F1 & ASR \\ \cmidrule(lr){2-3}  \cmidrule(lr){4-5}  \cmidrule(lr){6-7}  \cmidrule(lr){8-9}  \cmidrule(lr){10-11} \cmidrule(lr){12-13} \cmidrule(lr){14-15} \cmidrule(lr){16-17} \cmidrule(lr){18-19}\\
YOLOv3 (0\%) & 94.2 &  &  &  &  &  &  &  &  &  &  &  &  &  &  &  &  &  \\
  30\% & 82.2 &  22.8 & 80.1 & \textbf{28.9} & 81.9 & 21.1 &  78.7 & 20.2 & 79.7 & 22.1 & 82.7 & 17.3 & 86.0 & 18.4  & 77.4 & 26.4 & 84.8 & 20.2 \\
   40\% & 66.8 & 42.5 & 62.0 & 41.3 & 64.2 & 44.6 &  65.6 & 46.2 & 68.3 & 36.4 & 65.9 & 46.0 & 75.1 & 35.0 & 64.0 & 38.8 & 71.5 & \textbf{52.2} \\
  YOLOv10 (0\%) & 94.9 &  &  &  &  &  &  &  &  &  &  &  &  &  &  &  &  &  \\
   20\% & 88.8 &  9.2 & 90.1 & 9.1 & 89.5 & 8.4 & 90.1 & 7.1 & 87.9 & \textbf{10.0} & 92.2 & 6.4 & 91.4 & 6.8  & 92.2 & 6.2 & 90.8 & 5.2 \\
   30\% & 79.9 & \textbf{39.4} & 86.1 & 19.0 & 85.3 & 17.3 & 82.3 & 22.9 & 85.4 & 22.4 & 81.3 & 23.6 & 85.4 & 21.4 & 82.3 & 24.0 & 81.7 & 18.3 \\
   40\% & 64.6 & 42.2 & 72.8 & 43.4 & 66.1 & 44.1 & 69.2 & 43.1 & 65.4 &  41.0 & 69.1 & \textbf{49.5} & 76.7 & 37.1 & 72.5 & 44.7 & 66.3 & 39.4 \\ \bottomrule
\end{tabular}
\label{tab:compare_models}
\end{table}

The performance of the attack on the MilCivVeh dataset is limited. In the original BadDet publication, the attack achieves an ASR of 48.5\% (MS COCO) and 75.7\% (VOC 2007) when targeting the YOLOv3 model. This was achieved at a 30\% poisoning rate and with a patch size of 49 by 49 pixels. On MilCivVeh, at a 30\% poisoning rate, a maximum ASR of 28.9\% is reached for YOLOv3 with a patch size of 15 by 15 pixels. Significantly better results are reached with a higher poisoning rate of 40\% and a patch of 50 by 50 pixels: 52.2\% ASR. However, the performance of the model also severely degrades at that point, making it unlikely that the attack goes unnoticed. The minimum ASR that we see over all different settings is 17.3\%, which could still cause problems in an operational military setting. Thus, BadDet is not as efficient in the MilCivVeh dataset as in MS-COCO or VOC2007, we can still draw the careful conclusion that adversarial patch attacks could be a threat for object detectors in the military domain.

When targeting YOLOv10, the ASR does not exceed 50\%. Notably, the highest ASR for a poisoning rate of 30\% (ASR 39.4\%) is reached with the smallest patch size evaluated: 10 by 10 pixels. All other results are at least 15\% lower. For the higher poisoning rate of 40\% the ASR, though more stable across patch sizes, is the highest at 49.5\% at a patch size of 35 by 35, achieving similar performance to MS COCO in the original publication, though at the cost of a higher poisoning rate. For YOLOv10, the minimum ASR over the different settings is just 5.2\%, at the smallest poisoning rate. However, it should be taken into account that this is for a poisoning rate of only 20\%.

Overall, it stands out that the results show no clear trend on the relation between patch size and ASR. It seems that both a small and a large patch can lead to a high ASR and that the results are quite unstable in this regard: they differ strongly between the models, poisoning rates and patch sizes. This leaves the impression that choosing the right poisoning rate and patch size is a complex process that requires trial-and-error, a process a potential adversary would, presumably, not have access to and would be difficult to go unnoticed. 

Furthermore, when evaluating the results of the experiments across clean and poisoned data, it is noticeable that a high degree of misclassifications in the target class where of clean images. I.e., an image without adversarial patch would be missclassified by the object detection model. Often, the degree of misclassifications was similar or even higher than the ASR on the poisoned images. This suggests that, rather than learning to associate the patch with a specific class, the model (also) associates other features of the other classes with the target class, since it has been trained on 30\% mislabeled images. This high degree of misclassifications on clean test data is a logical consequence of such a high poisoning rate, but it makes it very difficult to differentiate between the impact of misclassifications due to the fact that images have been mislabeled in the poisoning process, and misclassifications as a result of the adversarial patch. To further investigate this, a lower poisoning rate of 20\% was added to the YOLOv10 experiments. However, this resulted in a poor ASR of maximum 10\%. 

All in all, the results are mixed: in some cases BadDet reaches an ASR of up to 52\%, a risk that surely is worth avoiding in the military context. However the degree to which this is caused by the adversarial patch rather than the misclassified training data is still an open question. Additionally, there is no clear trend to be found in the optimal combination of patch size, poisoning rate and target model that leads to a high ASR.



\subsection{Evaluating patch detection methods}
\label{res:detection}

\Cref{tab:compare_detection} shows the results of our initial set of experiments. Several datasets were poisoned using BadDet (GMA) with the HTBD patch of 25 by 25 pixels at a random location. The results of TRACE, taken from \cite{Zhang2025test} are also included. Then, the different anomaly and patch detection methods, together with the proposed AutoDetect method, were used to detect the poisoned images.

\begin{table}[t]
\centering
\caption{A comparison of the AUROC of the detection methods across datasets with random patch positions, with a constant patch size of 25x25 and the HTBD patch at a random location.}
    \begin{tabular}{llll} \toprule
    & MS COCO& MilCivVeh& VOC2007\\ \cmidrule(lr){2-4}
  AutoDetect (ours) & 0.965 & 0.941 &   0.970 \\
  PAD & 0.633 & 0.902  & 0.725 \\
  PaDIM  & 0.637 & 0.538 & 0.555 \\
  TRACE \footnotemark  & 0.924 & - & 0.939 \\  \bottomrule \end{tabular}
    \label{tab:compare_detection}
\end{table}
\footnotetext{These results on YOLO are taken from Zhang et al. \cite{Zhang2025test}.}
\paragraph{Anomaly detection methods} All the anomaly detection methods from the industrial inspection domain -- CFA, Efficient AD, PatchCore and PaDIM -- achieved subpar performance and resulted in long computation time and high memory usage. Specifically the methods CFA, EfficientAD and PatchCore performed subpar and achieved AUROC scores on COCO of $\sim 0.5$. We took this poor performance on MS COCO as a sufficient indicator that these methods, though they achieved success in the industrial anomaly detection domain, are not able to accurately detect adversarial patches in an object detection setting. For this reason, the experiments on MilCivVeh and VOC2007 were not completed for these methods and thus, the results are not depicted in \Cref{tab:compare_detection}. The method PaDIM did achieve results better than random, so this method was fully evaluated, although not with successful results. 
 
While we can reproduce the results from the original PatchCore work \citep{roth2022towards}, we observe that its predictions for our use case of real-world images with adversarial patches does not exceed a random choice baseline. We conjecture that the reason for this, is the inability of the Patchcore \textit{coreset} to capture the large variety of scenes and objects in our images. The evaluation settings used in \citep{roth2022towards} are industrial anomaly detection tasks containing a limited set of objects against simple backgrounds, such as in the MVTecAD dataset \cite{bergmann2019mvtec}. Such image sets may be effectively described with a limited set of vectors, while a large, complex image sets require prohibitively large \textit{coresets} to be captured effectively. As a result, a distance measurement from an unseen image to the insufficiently informed coreset is not meaningful.

We can view AutoDetect through the lens of the PatchCore method, in the sense that it too evaluates the proximity of a previously unseen image to the set of clean images. However, AutoDetect does not explicitly store a limited set of image representations, but instead uses a pre-trained model that captures the complex space of images. It uses this model to implicitly evaluate whether a new image is on the learned manifold of clean images by observing the reconstruction error for the unseen image. This approach requires orders of magnitude less memory and therefore scales well to our use case involving highly diverse images.

\paragraph{PAD} One method originally designed for adversarial patch localization and removal was tested: PAD. As can be seen in \Cref{tab:compare_detection}, this method, after the changes that were made as discussed in \Cref{sec:experimental_setup}, achieves good detection performance, especially on MilCivVeh. We do see that performance declines when the datasets become larger, and more diverse. The achieved AUROC for MS COCO is 0.633 and for VOC2007 0.725. Additionally, since PAD is meant for a more complex task than solely anomaly detection, namely anomaly localization and removal, it is very computationally intensive to run. This is why both the VOC2007 (25\%) and MS COCO (10\%) datasets were downsampled to generate results as shown in \Cref{tab:compare_detection}. Without downsampling, it would not have been possible to generate results within a week, while this could easily be done on comparable hardware with AutoDetect.

\paragraph{AutoDetect} our novel method achieved good detection performance on all three datasets -- especially compared to the other methods. Even though the underlying autoencoder was trained solely on MS COCO, the method also reached a high AUROC on the poisoned versions of MilCivVeh and VOC2007. Detection results are slightly lower for MilCivVeh which might be explained by the smaller dataset size, making the results less stable, and the fact that MilCivVeh captures a different domain (military) than MS COCO and VOC2007. 

It is relevant to note here that the Detector Cleanse, coined by Chan et al. \cite{chan2022baddet}, reached a maximum accuracy of 83\% \cite{chan2022baddet} on VOC2007, whereas our method achieved an accuracy of 95\%\footnote{The detection threshold was set at 0.95 to achieve these results} (AUROC 0.97) for that dataset. We achieve slightly better results than the current state-of-the-art patch detection method on the GMA attack: TRACE \cite{Zhang2025test}. This method, which focuses on detecting poison the trained model rather than the training data, achieved AUROC scores of 0.924 (YOLO), 0.846 (DETR) and 0.883 (Faster-RCNN) for MS COCO and AUROC's of 0.939 (YOLO), and 0.774 (DETR) for PASCAL VOC.

\section{Ablation studies}
\label{sec:ablation}  

To investigate the robustness of our AutoDetect method, we performed several ablation studies. First, we research the performance of AutoDetect given different adversarial patch sizes and patch types. Subsequently, we also look into the interplay between adversarial patch size and AutoDetect slice size. 


\subsection{Different patches and patch sizes}
\label{sec:abl_patches_sizes}

First, AutoDetect was tested on all three datasets for varying patch sizes and the three different patch types. Additionally to the HTBD patch used in the main research, we also have a checkerboard patch and a banana patch (\Cref{fig:patches}). The checkerboard is chosen since the original BadDet work uses a checkerboard patch, so this makes for a nice comparison. The banana is chosen since it is significantly different from the other two patches, but still very distinct. This could give a good indication whether the appearance of a patch has an impact on the detection performance.

\begin{figure}[t]
    \centering
    \includegraphics[width=0.5\linewidth]{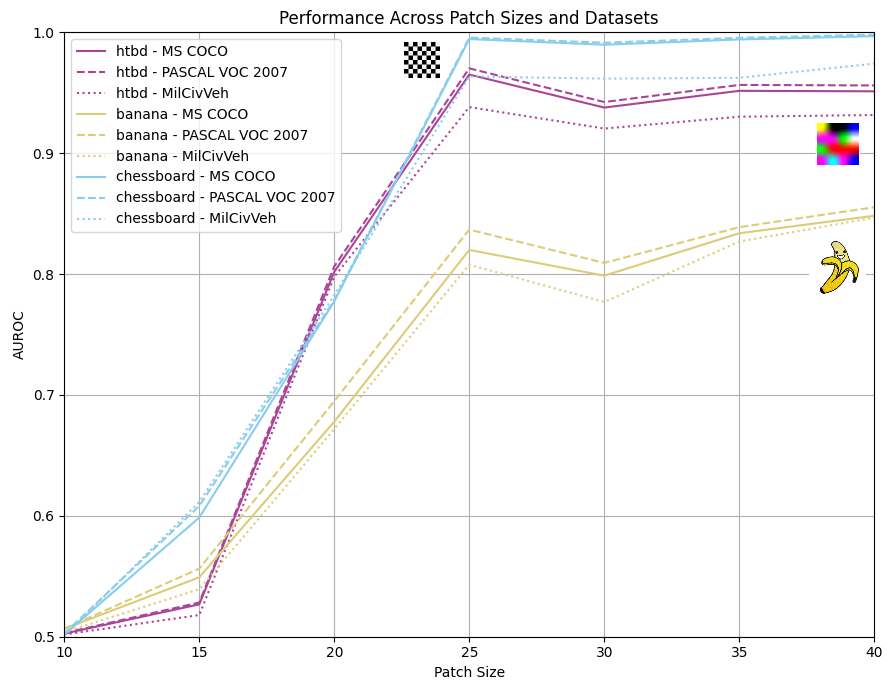}
    \caption{Performance measured by AUROC of AutoDetect across three datasets, three patches and varied patch size, with a constant AutoDetect slice size of 25x25px.}
    \label{fig:patches_sizes_datasets}
\end{figure}

The resulting AUROC scores can be found in \Cref{fig:patches_sizes_datasets}. Zooming in on the performance with the HTBD patch we note a similar trend across all three datasets: performance peaks when the adversarial patch size is equal to the AutoDetect slice size. We see a slight decrease in performance for a patch size of 30 by 30. with the performance increasing again after that. The performance for patches smaller than the slice size is relatively low. AutoDetect barely succeeds in detecting patches of size 15 by 15 or 10 by 10. It is questionable whether these patches even impact performance of a model (if they are able to poison it) enough to be relevant for detection. This weak performance on small adversarial patches can be explained by the fact that the mean slice loss plummets quite quickly once a relatively smaller part of the slice is covered by the patch. Using another metric to aggregate the slice loss instead of the mean, such as the max value or .875 percentile, could alleviate these issues. Experiments were conducted to evaluate these metrics, but they did not result in improved performance overall. Alternatively, when the patch size is larger, at least one slice is completely covered by the patch and thus the mean of that slice should still be high. 

A similar pattern in the relation between AUROC and patch size can be perceived for all three patches, with the notable difference that the banana patch is more difficult to detect across the board, with performance peaking at the largest patch sizes rather than the patch size equal to the slice size. The highest performance is reached for the chessboard patch which presents a stark contrast that is rarely seen in the real world and thus might cause a high loss. On the contrary, the banana patch has a white border around the figure, limiting contrast in a major part of the patch and thus limiting the loss in the patch as a whole. A notable conclusion to draw from this study, is that the patch choice does have an impact on the detection performance. This could mean that some patches are hard to detect, although we can imagine that his affects the performance of the attack as well.

When comparing performance across datasets, it is notable that for all patches and most patch sizes, AutoDetect achieves the lowest performance on poisoned MilCivVeh images. This might be caused by the fact that it is the most out-of-domain compared to MS COCO, on which the AutoDetect autoencoder is pretrained. VOC2007, though a different dataset, still has a lot of overlap with MS COCO in terms of scenes and objects featured. For MilCivVeh, this overlap is limited. Despite this, it is impressive that AutoDetect still consistently reaches a detection performance of AUROC $>0.9$ on the chessboard and htbd patches within MilCivVeh images. 

\subsection{Varying patch and slice size}
\label{sec:autodetect_analysis2}


The previous ablation study raises the question to what degree there is a mutual effect between the patch and slice size on the detection performance. For example, it could be that a smaller slice size results in improved results when detecting smaller patches. Thus, we conduct another study in which we vary both patch and slice size. To keep the experiments representative, but also concise, we only use the MS COCO dataset and the HTBD patch. 

\begin{table}[h]
\centering
\caption{Results (AUROC) of AutoDetect across varying slice and patch sizes, for MS COCO with patch HTBD, including the mean AUROC across all patch sizes for each slice. The highest performance per column is printed in bold. Rows denote the slice size, and columns denote the patch size.}
\begin{tabular}{lllllllll} \toprule
  \textbf{Slice \textbackslash{} Patch} &   \textbf{10} & \textbf{15} & \textbf{20} & \textbf{25} & \textbf{30} & \textbf{35} & \textbf{40}  & \textbf{Mean} \\ \cmidrule{2-9}
 \textbf{10} & \textbf{0.632} & 0.617 & 0.651 & 0.803 & 0.791 & 0.812 & 0.860  & 0.738 \\
  \textbf{15} & 0.508 & \textbf{0.831} & 0.823 & 0.853 & 0.884 & 0.926 & 0.927  & \textbf{0.822} \\
  \textbf{20} & 0.503 & 0.601 & \textbf{0.913} & 0.905 & 0.929 & 0.916 & 0.944  & 0.816 \\
\textbf{25} & 0.503 & 0.527  & 0.802 & \textbf{0.965} & 0.938 & 0.952 & 0.951  & 0.805 \\
  \textbf{30} & 0.503 & 0.515 & 0.653  & 0.914 & \textbf{0.976} & 0.956 & 0.962  & 0.783  \\
  \textbf{35} & 0.502 & 0.511 & 0.572 & 0.843 & 0.947 & \textbf{0.982} & 0.968  & 0.761 \\
  \textbf{40} & 0.502 & 0.509 & 0.544 & 0.755 & 0.901  & 0.964 & \textbf{0.986} & 0.737    \\
\textbf{Mean	}& 0.522 & 0.587	& 0.708	& 0.863	& 0.909	& 0.930	& \textbf{0.942} \\ \bottomrule
\end{tabular}
\label{tab:patch_slice_size}
\end{table}

\Cref{tab:patch_slice_size} shows the results of AutoDetect across varying slice and patch sizes. It is clear to see that there is indeed a relation between patch and slice size: the performance for a given patch size is always highest when the slice size is equal to that patch size. Additionally, the smaller the patch, the lower the detection performance. A patch of size 10 by 10 is almost impossible to detect, even with a slice size of 10 by 10. This might be caused by the details of the patch itself being mostly lost at this patch size. The highest average detection performance is reached with a slice size of 15 by 15, lower than our selected size of 25 by 25. However, this is partially due to high performance on smaller patch sizes, which might be less relevant to detect due to their limited ability to poison the model. Based on this evaluation, future users might select a different slice size depending on the expected patch size in their use case.

\begin{figure}[t]
    \centering
    \includegraphics[width=0.5\linewidth]{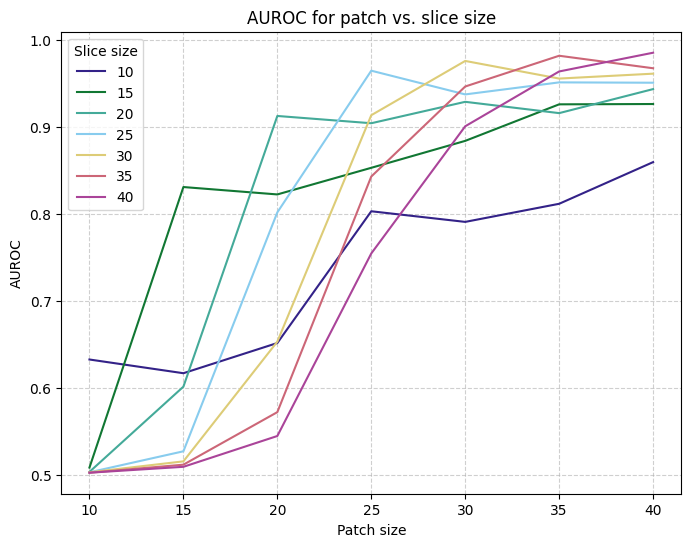}
    \caption{Results (AUROC) of AutoDetect across varying slice and patch sizes, for MS COCO with patch HTBD, including the mean AUROC across all patch sizes for each slice, visualized.}
    \label{fig:patch_slice_auroc}
\end{figure}

Additionally, \Cref{fig:patch_slice_auroc} visualizes the data in \Cref{tab:patch_slice_size} to make the relation between patch and slice size more visible: most lines peak when the patch size is equal to the slice size, and performance plummets after a patch size of 20 by 20. The slice size of 10 by 10 performs poorly across almost all patch sizes. A potential explanation for this is the fact that such a small slice might be very sensitive to noise. Another notable trend, which was also noted in the previous section, is that performance for larger patch sizes relatively good across slice sizes, sometimes even higher than for an exact match in size.






\section{Discussion}
\label{sec:discussion}  
This paper aimed to investigate whether military object detection systems are vulnerable to poisoning attacks, whether existing patch, or anomaly, detection methods could be applied to the domain, and whether our novel proposed method, AutoDetect, could successfully detect poisoned samples. 

Based on the results, we can draw the conclusion that military object detection systems are indeed vulnerable to the BadDet attack. However, based on the high poisoning rate required, in combination with the seeming invariance to patch size, it seems that it is the mislabeled data rather than the adversarial patch itself that causes misclassifications. The mislabeling of 20 percent or more of the data required to successfully apply BadDet, could be detected during manual inspection of the data or when evaluating performance on a correctly labeled test set. This suggests that, in its current form, object detection poisoning attacks are underdeveloped. In the field of poisoning attacks targeting classification models in the image domain, we do see very stealthy attacks that require smaller poisoning rates, e.g., clean-label attacks featuring perturbations \cite{aghakani2021bullseye, zeng2023narcissus, Jiang2023color}. If these attacks were to be transferred to the object detection task, this could cause major risks, much greater than the ones caused by BadDet, since they would require both a lower poisoning rate and are more difficult to detect through manual inspection. 


When using existing anomaly detection methods to detect poisoned images, we encountered various problems. In hindsight, these methods might not have been an appropriate match for our problem, since most of them are built on the assumption that all training images are very similar: in the industrial inspection benchmarking dataset MVTec-AD \cite{bergmann2019mvtec}, all images of the same class are very similar. The only small changes are the anomalies in the form of product defects. This calls for a different detection approach than in our setting, where the dataset has more variety within classes. PAD, a method originally designed for adversarial patch localization and removal \cite{Jing_2024_CVPR}, achieved a reasonable performance on MilCivVeh, but performed poorly on MS COCO, the largest dataset. An advantage of PAD is the fact that it does not need a clean subset of images before it is able to detect the adversarial patches, which might make it a good choice when it is impossible to gather clean images.  All AD methods and PAD have shown to be very memory intensive and thus limited in their practical application: if it is so computationally expensive to run them, they might not be suitable for use in a military setting. Thus, we can conclude that AD methods from the visual industrial inspection domain are no match for adversarial patch detection in the military domain.

In this work we introduced AutoDetect, a novel poisoning detection method based on an autoencoder (pretrained on MS COCO). The method demonstrated the ability to detect poisoned samples in all three datasets with a high AUROC (0.94+), resulting in better or similar performance to the current state-of-the-art. AutoDetect was further evaluated for more patches and patch sizes, showing that it was able to detect patches primarily with a high contrast, across datasets. Also, the relation between patch and slice size was clear and thus it seems important to choose a slice size in line with the expected patch size. 

\subsection{Limitations}
\label{sec:limitations}

One issue that could be pointed out for the AutoDetect method is that it needs a clean set of data from the target domain in order to fit the distribution of losses. One might argue that this shifts the issue of not knowing whether a dataset is poisoned to an earlier step in the process. However, we would like to point out that the distribution for the detection of poisoned samples in MilCivVeh was only fitted to a subset of 99 (validation) images. Manually checking this smaller sample of data still saves considerable time compared to manually checking the full training set. Another solution to this issue could be to employ a method to gather a clean subset of the data, such as Meta-Sift \cite{zeng2023meta}. On the other hand, the autoencoder part of AutoDetect  does not need to be trained on a domain-specific dataset, as was highlighted earlier, and was solely trained on MS COCO. This does allow for more flexibility when deploying AutoDetect.

One limitation of the metrics we used to measure AutoDetect's performance that should be mentioned, is that AUROC is threshold independent, whereas AutoDetect is not. We use AUROC because it is representative of the full potential of the method, though it might not be representative of its actual performance in the real world -- after all, the user of AutoDetect has to manually set a threshold for detection. However, the ideal threshold also depends on the use case and the risk appetite of the user. Thus, we found it most representative to report AUROC and occasionally add the accuracy when comparing to other studies. 

In terms of the data used in this work, the military dataset MilCivVeh could be seen as limited. MilCivVeh is a custom dataset which is relatively small, limited in the number of classes it features and, due to being based on public data, limited in its representation of operational military data. It also includes mostly ground-to-ground perspectives, where a air-to-ground might be more useful in many military use case. A larger, more representative dataset would be necessary before making definitive statements about the applicability of both adversarial attacks and our AutoDetect defence method in the military domain. However, we do think that MilCivVeh is useful to give a first insight into the threat of poisoning attacks in the military object detection domain. 

Furthermore, the method of patch insertion into the dataset, though taken from the original BadDet publication \cite{chan2022baddet}, might not be the most representative of a real-world scenario, specifically in the military domain. One might expect that rather than being inserted digitally after collection, patches might be brought physically into an operational setting. Right now, it is not clear yet whether physically adding patches to scenes solicits the same behavior both in terms of Attack Success Rate and in terms of ability to detect these patches using AutoDetect. One could imagine that training data might have been poisoned digitally, but in the field (at test time) it is to be expected that an adversary uses physical patches to evade the object detector.  

The last analysis in \Cref{sec:autodetect_analysis2} is conducted only on MS COCO, which limits the robustness of the conclusions presented there. MS COCO was selected because it is the largest and most popular dataset and the data set the autoencoder is fitted on. The fact that the autoencoder underlying AutoDetect was fitted to this dataset might seem like an unfair advantage, however, we stress that even though the autoencoder was trained on this dataset, the results on MS COCO were not the highest in any of the other experiments. Additionally, HTBD was selected because it also had neither the lowest nor the highest rate of detections in our experiments, and is considered most representative of the adversarial patches used in this domain. 


\subsection{Future work}
\label{sec:future_work}
To further explore the risks and opportunities for the detection of adversarial patches in the military domain, large, representative datasets are needed in the military domain. Specifically, datasets that incorporate adversarial patches in an operational physical setting could significantly boost the field of adversarial patch detection. We also understand that creating a dataset with physical patches, especially in the military domain, is time consuming and difficult. However, testing against an existing dataset with physical patches, even if in a different domain, might be a good first step to explore the usability of AutoDetect in an operational military setting \cite{braunegg2020apricot}. This could also give a more clear indication whether poisoning attacks are a realistic threat to object detection models in the military domain.

Lastly, there are still some aspects of AutoDetect and BadDet that could be further investigated. A variety of three patches was selected to test the efficiency of BadDet and AutoDetect. An even more diverse set of patches could be part of future research, to test the effect of different patches on both attack performance and detection performance. Additionally, an evaluation of the actual effectiveness of poisoning a model with these patches and a comparison to performance when using AutoDetect to remove poisoned images from the dataset would be very useful to show the usefulness of the method. Finally, it would be interesting to see if AutoDetect is also an effective countermeasure against different patch-based attacks than BadDet's GMA attack, such as the works used for evaluation by the authors of TRACE \cite{Zhang2025test}. It might also be worthwhile to test the performance of AutoDetect on poisoning attacks that are not patch-based, but we assume that this will not be very successful, since AutoDetect has been designed to find adversarial patches.

\newpage
\bibliography{refs} 
\bibliographystyle{spiebib} 

\end{document}